\documentclass[letterpaper]{article} 
\usepackage[draft]{aaai2027}  
\usepackage[hyphens]{url}  
\usepackage{graphicx} 
\usepackage{natbib}  
\usepackage{caption} 
\usepackage{algorithm}
\usepackage{algorithmic}

\usepackage{newfloat}
\usepackage{listings}
\DeclareCaptionStyle{ruled}{labelfont=normalfont,labelsep=colon,strut=off} 
\floatstyle{ruled}
\newfloat{listing}{tb}{lst}{}
\floatname{listing}{Listing}

\usepackage{booktabs}
\usepackage{multirow}

\usepackage{amsmath}
\usepackage{amssymb}

\newcommand{\best}[1]{\textbf{#1}}
\newcommand{\NA}{--}

\title{SpecPrefetch: Parameter-Efficient Expert Prefetching for Sparse MoE Foundation Models}
\author{
    Jinwei Kong\textsuperscript{\rm 1,2,*},
    Runqi Meng\textsuperscript{\rm 2,*},
    Xiaolong Zhong\textsuperscript{\rm 1},
    Fanyi Wang\textsuperscript{\rm 1},
    Wentao Qiu\textsuperscript{\rm 1,3},\\
    Haotian Hu\textsuperscript{\rm 1},
    Yongjian Zhou\textsuperscript{\rm 4},
    Zhenhua Ge\textsuperscript{\rm 1,\ensuremath{\dagger}}
}
\affiliations{
    \textsuperscript{\rm 1}StepX,
    \textsuperscript{\rm 2}ShanghaiTech University,
    \textsuperscript{\rm 3}Xiamen University,
    \textsuperscript{\rm 4}Chongqing University\\
    \textsuperscript{*}Equal contribution;
    \textsuperscript{\ensuremath{\dagger}}Corresponding author.\quad
    \\
    kongjw2023@shanghaitech.edu.cn,\quad gezhenhua@ustc.edu
}

\begin{document}

\maketitle

\begin{abstract}

Sparse Mixture of Experts (MoE) models enable scalable foundation models through conditional computation, but their large expert pools pose significant memory challenges for deployment. Expert offloading reduces accelerator memory usage by storing inactive experts in host memory or storage, yet introduces additional transfer latency because required experts are only identified after native routing. This reactive loading places expert transfer on the inference critical path. We propose SpecPrefetch, a parameter-efficient and router-preserving framework for expert prefetching in offloaded MoE inference. SpecPrefetch employs lightweight layer-specific adapters to predict next-layer expert priorities solely for asynchronous transfer, while preserving the frozen native router for final expert selection. To avoid excessive transfers beyond the achievable overlap capacity, SpecPrefetch introduces a window-aware prefetch budgeting policy based on the profiled computation and transfer overlap window. This design improves expert availability without modifying pretrained routing behavior. Experiments on Qwen3-VL-30B-A3B and DeepSeek-VL2-Tiny demonstrate that SpecPrefetch achieves superior next-layer expert coverage across diverse workloads with substantially fewer trainable parameters than learned predictor baselines. Real-device evaluation on Snapdragon 8 Elite further shows up to 20\% decoding throughput improvement under storage-constrained settings over a compute-optimized offloading runtime, validating the effectiveness of prediction-guided and budget-aware expert prefetching for resource-constrained MoE deployment. Code and weights will be released upon acceptance.
\end{abstract}

\section{Introduction}

\begin{figure}[!t]
  \centering
  \includegraphics[width=\columnwidth]{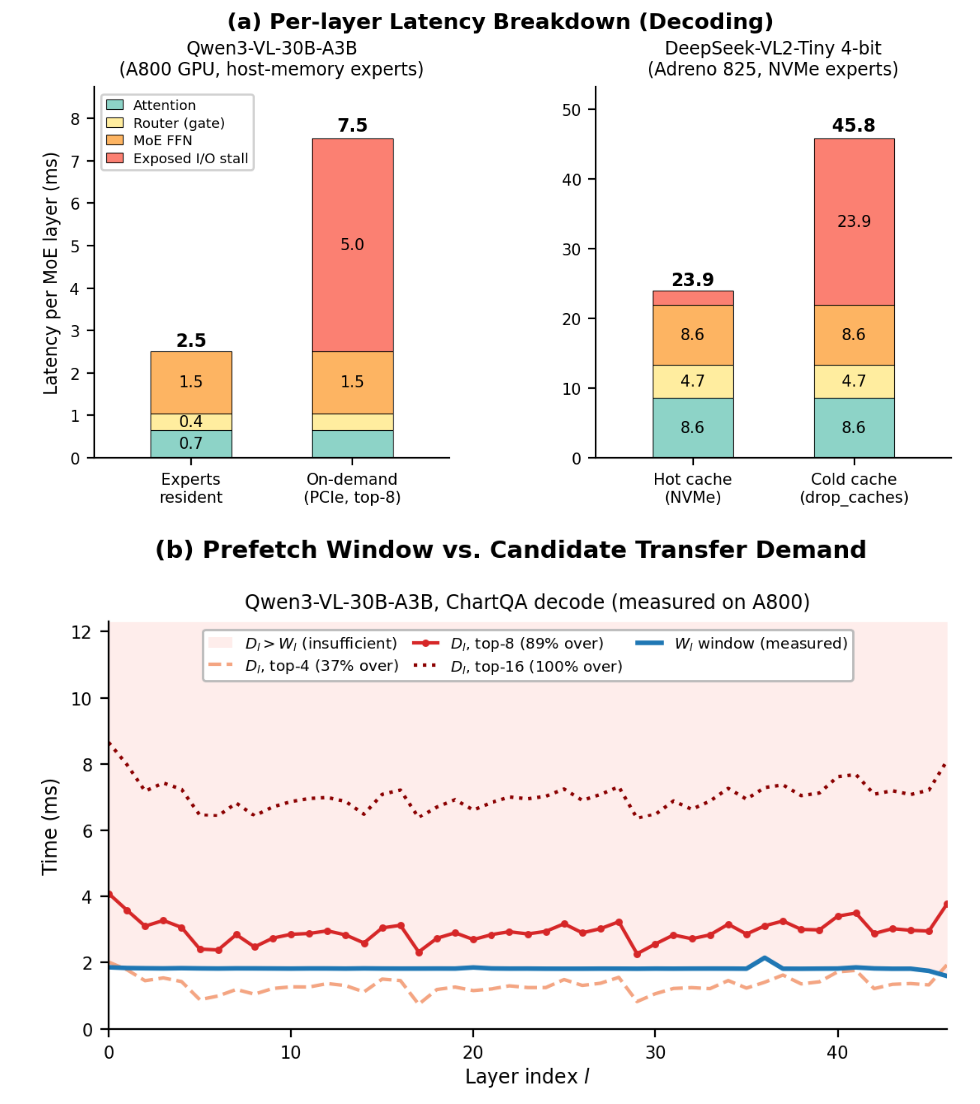}
  \caption{
Motivation for window-aware expert prefetching.
(a) Per layer decoding latency under resident and on-demand expert loading. Exposed expert loading accounts for 67\% of per layer latency on the GPU and 52\% under mobile cold cache, while MoE computation provides an opportunity for transfer overlap.
(b) Measured prefetch window and candidate transfer demand on Qwen3-VL-30B-A3B. Curves report per layer averages, while percentages denote the fraction of event level demands exceeding their corresponding prefetch windows. The window supports about two transfers, while top-4 and top-8 exceed it in 37\% and 89\% of events.
}
  \label{fig:intro_0}
\end{figure}

Recent progress in large language models (LLMs) and vision language models (VLMs) has been driven by continued growth in model capacity, training data, and computation. However, dense scaling incurs substantial inference costs in memory, computation, and latency. Mixture of Experts (MoE) architectures provide a more efficient scaling paradigm by retaining a large parameter capacity while activating only a small subset of experts for each token. Existing MoE models adopt either routed expert pools or hybrid strategy with shared and routed experts~\citep{qwen_moe, jiang2024mixtral, dai2024deepseekmoe}. Although sparse activation reduces per token computation, inference still requires access to the complete expert pool. Consequently, the large expert footprint becomes a major obstacle to deploying MoE models on commodity GPUs and resource constrained edge devices. Expert offloading reduces memory pressure by storing inactive experts in CPU memory or host storage and transferring selected experts during inference.

Nevertheless, expert offloading introduces substantial data movement because the required experts remain unknown until the native router evaluates the corresponding hidden representations. Consequently, conventional on-demand loading places expert transfer on the inference critical path and stalls the accelerator before MoE execution. Figure~\ref{fig:intro_0}(a) quantifies the resulting stall and further shows that MoE FFN computation accounts for a substantial fraction of per layer latency. Meanwhile, the relatively long FFN stage provides an opportunity to overlap next-layer expert transfer with ongoing computation. Exploiting the overlap window requires expert demand to be predicted before the target router is executed. However, the available window may be insufficient to transfer all predicted candidates. As shown in Figure~\ref{fig:intro_0}(b), candidate transfer demand exceeds the available prefetch capacity in multiple layers. Although enlarging the candidate set can improve expert coverage, redundant transfers consume limited bandwidth and memory. Therefore, efficient offloaded MoE inference requires both accurate next-layer expert prediction and a bounded prefetch budget selected from the profiled computation and transfer overlap window.

\begin{figure}[t]
  \centering
  \includegraphics[width=\columnwidth]{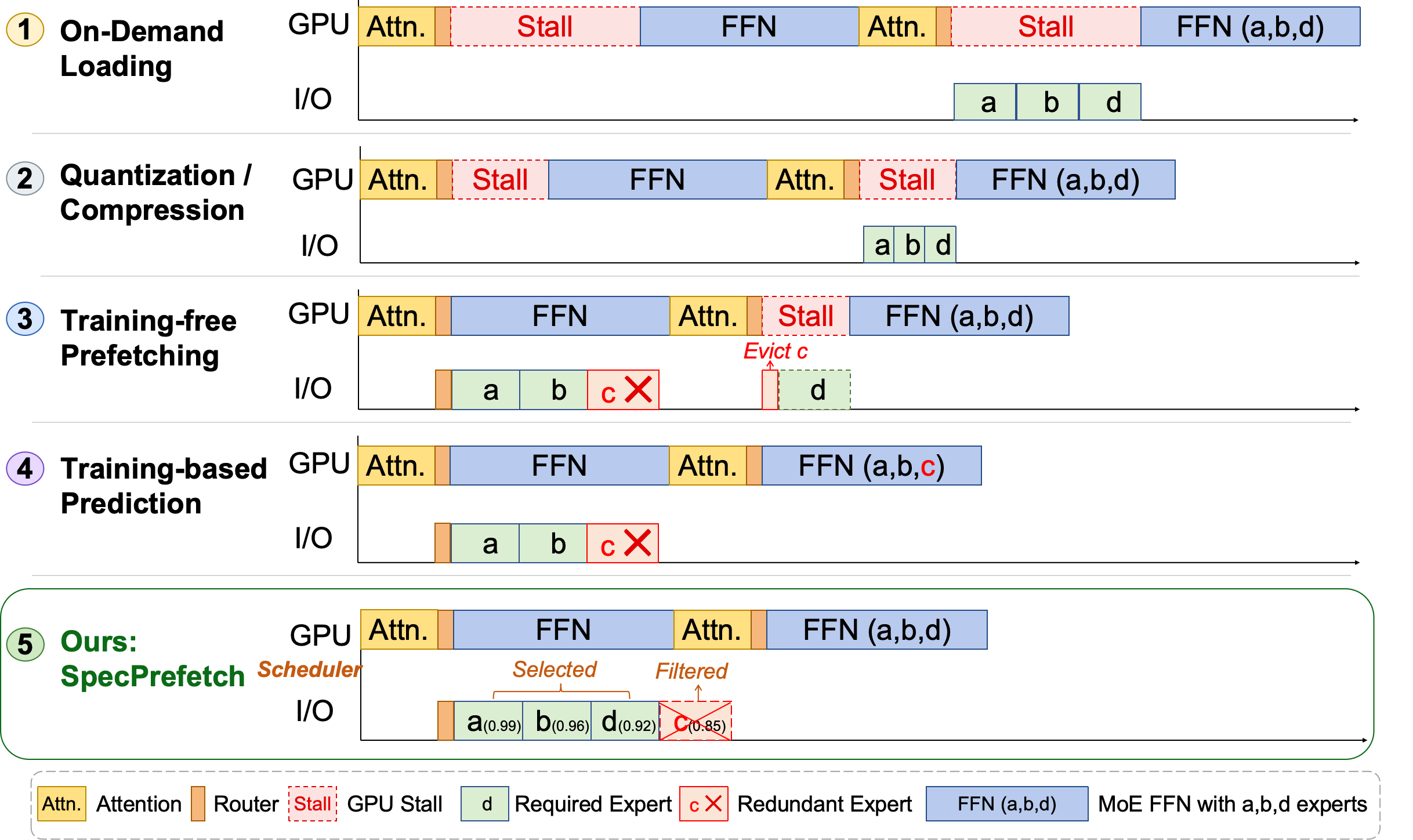}
  \caption{
  Comparison of expert-loading timelines in offloaded MoE inference.
  On-demand execution exposes expert-transfer latency after native routing.
  SpecPrefetch predicts next-layer expert candidates in advance and overlaps
  their transfer with ongoing computation, while the native router remains
  authoritative for final expert execution.
  }
  \label{fig:overview}
\end{figure}


Existing approaches address only individual aspects of the expert transfer bottleneck, as summarized in Figure~\ref{fig:overview}. Quantization and compression reduce expert size and transfer latency, but expert demand remains unavailable before native routing~\citep{moepic2025}. Training free methods, such as FATE, exploit cross layer routing correlations with negligible prediction overhead~\citep{fang2025fate}. However, limited prediction accuracy often requires a larger candidate set, thereby increasing redundant transfers and cache pressure. Training based methods improve predictive capacity through auxiliary predictors or pre gating mechanisms~\citep{hwang2024pregated,chen2025spmoe}, but may introduce additional latency and memory overhead or interfere with native routing decisions. More importantly, prediction accuracy alone does not determine prefetching effectiveness. A predicted expert contributes to latency reduction only when its transfer overlaps useful computation, whereas incorrect predictions consume limited bandwidth and memory resources. Effective expert prefetching must balance expert coverage, prediction overhead, and a window-derived prefetch budget, while preserving native routing.

To address these challenges, we propose SpecPrefetch, a parameter-efficient and router preserving framework for expert prefetching in offloaded MoE inference. SpecPrefetch employs lightweight layer-specific adapters to predict next-layer expert priorities and initiate asynchronous transfers, while retaining the native router for final expert selection. Moreover, a window-aware budgeting policy limits priority-based asynchronous prefetching according to a maximum budget selected from the profiled overlap window. By separating transfer prediction from execution routing, SpecPrefetch improves expert readiness without altering the routing behavior of the pretrained MoE model.

We evaluate SpecPrefetch at both the model and system levels. Model level experiments on Qwen3-VL-30B-A3B and DeepSeek-VL2-Tiny demonstrate accurate next-layer expert prediction and high native expert coverage with limited trainable overhead across language and multimodal workloads. Furthermore, real device deployment on a Snapdragon 8 Elite platform confirms that the improved expert readiness translates into practical acceleration, yielding up to a 20\% improvement in decoding throughput under storage constrained offloading conditions.

Our main contributions are as follows.

\begin{itemize}

    \item We formulate expert prefetching as a router preserving transfer problem that jointly considers expert coverage, prediction overhead, and the prefetch capacity provided by the computation and transfer overlap window.

    \item We propose SpecPrefetch, which combines parameter-efficient next-layer expert prediction with window-aware prefetch budgeting while preserving native routing for execution.

    \item We evaluate representative MoE models and deploy on a real mobile platform, demonstrating accurate expert prediction with low overhead and up to a 20\% improvement in decoding throughput.

\end{itemize}

\begin{figure*}[t]
  \centering
  \includegraphics[width=0.9\textwidth]{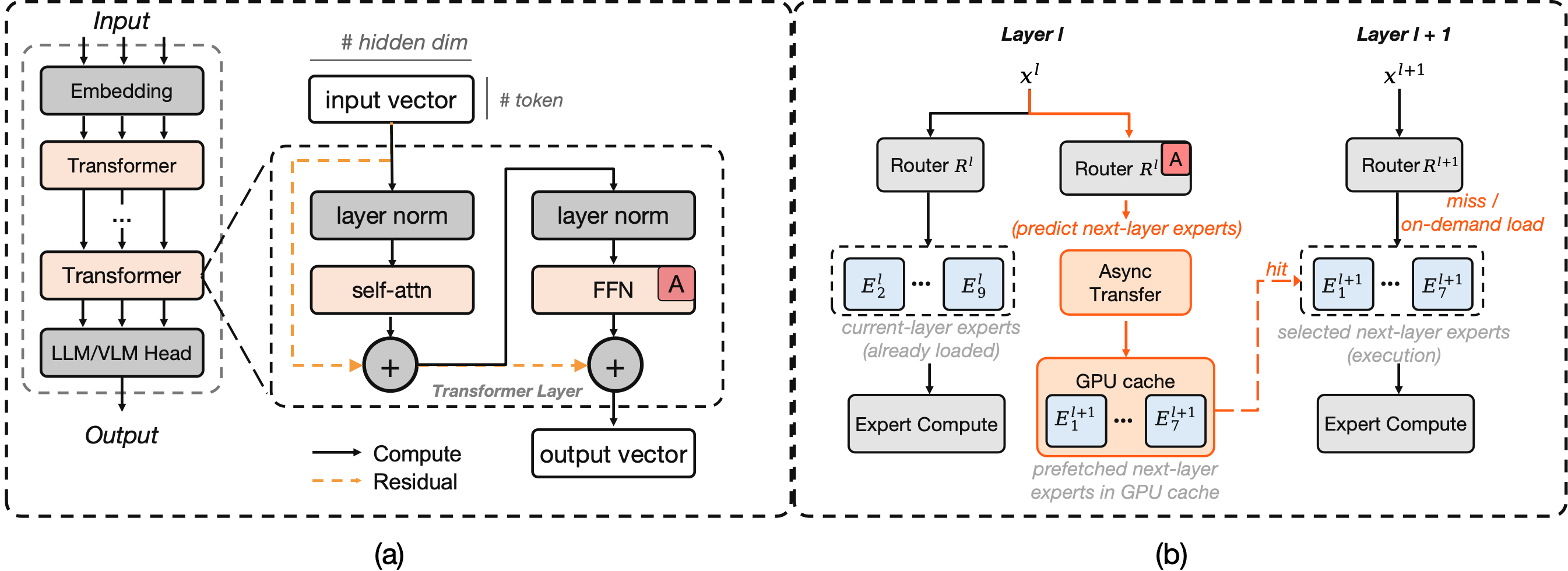}
\caption{
Overview of SpecPrefetch.
(a) A lightweight adapter is attached alongside each MoE block.
(b) During execution at layer \(l\), the adapter predicts next-layer expert priorities and asynchronously prefetches candidate experts, while the native router \(R^{l+1}\) remains unchanged for final expert selection.
}
  \label{fig:method}
\end{figure*}

\section{Related Work}
\subsection{Sparse MoE Foundation Models}

Mixture of Experts (MoE) scales foundation models through conditional computation. In sparse MoE layers, a router activates only a small subset of experts for each token, increasing total parameter capacity while limiting activated computation~\citep{shazeer2017outrageously}. Early systems such as GShard, Switch Transformer, and GLaM show that sparse routing improves model capacity and language modeling performance with controlled computation cost, while introducing challenges in routing stability, load balancing, communication, and expert specialization~\citep{lepikhin2020gshard,fedus2022switch,du2022glam}.

Recent MoE models further diversify expert architecture. Mixtral uses a pure routed expert pool~\citep{jiang2024mixtral}, DeepSeekMoE separates shared and routed experts~\citep{dai2024deepseekmoe}, and sparse upcycling initializes MoE models from dense checkpoints to reduce training cost~\citep{komatsuzaki2023sparse}. These designs differ in how expert capacity is constructed, shared, and routed, which directly affects offloading and expert prefetching.

\subsection{Offloaded MoE Inference and Expert Prefetching}
Expert offloading enables large sparse MoE models to operate under limited accelerator memory by storing inactive experts in CPU memory or host storage. Existing systems primarily optimize expert placement and data movement through caching, heterogeneous memory management, quantization, compression, expert splitting, and asynchronous transfer~\citep{semoe2022,moeinfinity2025,expertflow2024,layerscope2024,klotski2025,moepic2025,hobbit2024,finemoe2026}. In parallel, scheduling and prefetching methods overlap expert transfer with computation through predictive scheduling, token rebatching, pipeline execution, and expert aware memory organization~\citep{expertflow2024,klotski2025,layerscope2024,moeeras2024,dsmoe2024}. FATE further exploits adjacent layer routing correlations for training free expert prefetching on edge devices~\citep{fang2025fate}. However, many existing methods either optimize data movement after expert demand becomes available or rely on stable routing regularity and throughput oriented execution mechanisms.

Learned prefetching methods predict future expert activations through auxiliary predictors, pre gating mechanisms, or parameter-efficient adaptation~\citep{hwang2024pregated,chen2025spmoe,moebeyond2025,preattention2025}, while speculative generation creates additional transfer windows through draft and verification stages~\citep{leviathan2023speculative,eagle2024,moespeq2025}. In contrast, SpecPrefetch targets standard offloaded MoE inference and predicts next-layer candidates solely for asynchronous transfer. The native router remains responsible for final top-\(K\) expert selection, allowing SpecPrefetch to advance expert availability without altering the routing behavior of the pretrained model.

\section{Method}

As illustrated in Figure~\ref{fig:method}, SpecPrefetch separates expert prediction for transfer from native routing for execution. Section~\ref{sec:problem_formulation} formulates router preserving prefetching under a finite overlap window. Section~\ref{sec:expert_prediction} introduces a lightweight adapter that estimates next-layer expert priorities. Section~\ref{sec:window_aware_budgeting} uses the predicted priorities and a profiled maximum prefetch budget to initiate asynchronous transfers. Finally, Section~\ref{sec:training_objective} presents the training objective for the prediction adapters.

\subsection{Problem Formulation}
\label{sec:problem_formulation}

We formulate SpecPrefetch as an early expert transfer problem under a finite overlap window while preserving the native routing policy. Consider an MoE model with \(L\) MoE layers and \(n\) routed experts per layer. Let
\(\mathbf{x}_{b,s}^{l}\in\mathbb{R}^{D}\)
denote the router input of token \((b,s)\) at layer \(l\), and let \(\Omega\) denote the set of valid tokens in the current batch. The native router \(R^l\) selects \(K\) experts for each token:

\[
\mathcal{T}_{b,s}^{l}
=
\operatorname{Top}_{K}
\left(
R^l\left(\mathbf{x}_{b,s}^{l}\right)
\right).
\]

The corresponding batch-level native expert demand is

\[
\mathcal{T}_{\mathrm{batch}}^{l}
=
\bigcup_{(b,s)\in\Omega}
\mathcal{T}_{b,s}^{l}.
\]

The set \(\mathcal{T}_{\mathrm{batch}}^{l}\) contains all experts required by the current batch at layer \(l\) and is determined exclusively by the native router. Before routing is performed at layer \(l+1\), however, \(\mathcal{T}_{\mathrm{batch}}^{l+1}\) remains unavailable. For each \(l=1,\ldots,L-1\), SpecPrefetch therefore estimates an expert priority vector
\(\mathbf{s}^{l+1}\in\mathbb{R}^{n}\)
from the router inputs available at layer \(l\). The prediction guides only early transfer, while the native router remains responsible for final expert selection.

Let \(t_l^{\mathrm{pred}}\) denote the time at which the prediction result at layer \(l\) becomes available, and let \(t_{l+1}^{\mathrm{exec}}\) denote the time at which routed expert computation begins at layer \(l+1\). The available prefetch overlap window is

\[
W_l
=
t_{l+1}^{\mathrm{exec}}
-
t_l^{\mathrm{pred}}.
\]

Within \(W_l\), SpecPrefetch issues priority-based asynchronous transfers under a profiled maximum prefetch budget. Redundant transfers consume limited transfer capacity, while missed experts require on-demand loading. Since the native router remains unchanged, prediction errors affect inference efficiency rather than model outputs.

\subsection{Expert Prediction Adapter}
\label{sec:expert_prediction}

SpecPrefetch attaches an external low rank adapter to each adjacent pair of MoE layers. Given the router input
\(\mathbf{x}_{b,s}^{l}\)
at layer \(l\), the adapter predicts the expert logits of layer \(l+1\) as

\[
\hat{\mathbf{z}}_{b,s}^{l+1}
=
B^l A^l \mathbf{x}_{b,s}^{l}.
\]

Here,

\[
A^l\in\mathbb{R}^{r\times D},
\qquad
B^l\in\mathbb{R}^{n\times r},
\]

and \(r\ll D\) denotes the adapter rank. The corresponding token level expert distribution is

\[
\hat{\boldsymbol{\pi}}_{b,s}^{l+1}
=
\operatorname{softmax}
\left(
\hat{\mathbf{z}}_{b,s}^{l+1}
\right).
\]

Because expert transfers are scheduled at the batch level, token level predictions are aggregated into expert priorities,

\[
s_e^{l+1}
=
\sum_{(b,s)\in\Omega}
\hat{\pi}_{b,s,e}^{l+1},
\qquad
e=1,\ldots,n.
\]

The resulting priority vector is

\[
\mathbf{s}^{l+1}
=
\left[
s_{1}^{l+1},
\ldots,
s_{n}^{l+1}
\right].
\]

The score \(s_e^{l+1}\) represents the total predicted routing mass assigned to expert \(e\) and therefore prioritizes experts expected to serve more tokens in the current batch. For model-level evaluation with \(M\) predicted candidates, the candidate set is

\[
\mathcal{C}_{M}^{l+1}
=
\operatorname{Top}_{M}
\left(
\mathbf{s}^{l+1}
\right).
\]

Assuming uniform hidden dimensions and expert counts across layers, adapters for all adjacent MoE layer pairs introduce

\[
(L-1)r(D+n)
\]
trainable parameters.

\subsection{Window-Aware Prefetch Budgeting}
\label{sec:window_aware_budgeting}

The predicted priority vector
\(\mathbf{s}^{l+1}\)
ranks the experts at layer \(l+1\) according to their expected routing demand. Since expert transfer can reduce latency only when it overlaps with the computation preceding the target MoE layer, the number of prefetched experts must be constrained by the available overlap window.

Let
\(t_l^{\mathrm{pred}}\)
denote the time at which the prediction becomes available and
\(t_{l+1}^{\mathrm{exec}}\)
denote the start time of routed expert computation at layer \(l+1\). The available prefetch window is

\[
W_l
=
t_{l+1}^{\mathrm{exec}}
-
t_l^{\mathrm{pred}}.
\]

We profile the computation window and expert transfer latency on the target platform to determine a maximum prefetch budget
\(M_{\max}\).
This budget limits the number of transfer requests issued for each target MoE layer and prevents excessive prefetching from introducing redundant I/O and cache pressure.

At runtime, let
\(\mathcal{A}^{l+1}(t)\)
and
\(\mathcal{U}^{l+1}(t)\)
denote the experts already available and currently being transferred, respectively. These experts are excluded from additional transfer requests. The remaining experts are ranked according to
\(s_e^{l+1}\),
and the highest-ranked candidates are selected as

\[
\mathcal{P}^{l+1}(t)
=
\operatorname{Top}_{M_{\max}}
\left(
\left\{
s_e^{l+1}
\mid
e
\notin
\mathcal{A}^{l+1}(t)
\cup
\mathcal{U}^{l+1}(t)
\right\}
\right).
\]

The selected experts are asynchronously transferred in descending order of predicted priority. The actual number of issued transfers
\(m_l\)
satisfies

\[
m_l
=
\left|
\mathcal{P}^{l+1}(t)
\right|
\le
M_{\max}.
\]

At layer \(l+1\), the native router independently determines the experts used for execution. Any required expert that is not available is loaded on demand. Therefore, the prefetch budget affects only expert availability and transfer overhead, while the native routing policy and model outputs remain unchanged.

\subsection{Training Objective}
\label{sec:training_objective}

The frozen native router at layer \(l+1\) provides the target routing distribution

\[
\boldsymbol{\pi}_{b,s}^{l+1}
=
\operatorname{stopgrad}
\left[
\operatorname{softmax}
\left(
R^{l+1}
\left(
\mathbf{x}_{b,s}^{l+1}
\right)
\right)
\right].
\]

The prediction adapters are optimized by minimizing

\[
\mathcal{L}_{\mathrm{pred}}
=
\frac{1}{L-1}
\sum_{l=1}^{L-1}
\frac{1}{|\Omega|}
\sum_{(b,s)\in\Omega}
\operatorname{KL}
\left(
\boldsymbol{\pi}_{b,s}^{l+1}
\|
\hat{\boldsymbol{\pi}}_{b,s}^{l+1}
\right).
\]

Distilling the complete routing distribution preserves the relative expert ordering required for priority based prefetching. Only the adapter parameters
\(\{A^l,B^l\}_{l=1}^{L-1}\)
are optimized, while the pretrained MoE model remains frozen.

\begin{table*}[!t]
\centering
\caption{
Next-layer expert coverage of SpecPrefetch and baseline
predictors across LLM and VLM workloads.
The evaluated candidate counts represent limited, native, and
expanded prediction settings for each model.
}
\label{tab:prediction_next1_comparison}
\scriptsize
\resizebox{0.8\textwidth}{!}{
\begin{tabular}{lllcccccc}
\toprule
\textbf{Category}
& \textbf{Benchmark}
& \textbf{Method}
& \multicolumn{3}{c}{\textbf{Qwen3-VL-30B-A3B}}
& \multicolumn{3}{c}{\textbf{DeepSeek-VL2-Tiny}} \\
\cmidrule(lr){4-6} \cmidrule(lr){7-9}
&
&
& R@4$^{\dagger}$ & R@8 & R@10
& R@3$^{\dagger}$ & R@6 & R@8 \\
\midrule

\multirow{8}{*}{\shortstack[c]{\textbf{LLM}\\\textbf{Workloads}}}
& \textbf{GSM8K}
& FATE
& 47.51 & 81.40 & 88.38
& \best{48.62} & \best{84.66} & \best{93.07} \\
& & Draft Model
& \NA & \NA & \NA
& 46.52 & 78.57 & 87.56 \\
& & ProMoE
& 48.45 & 82.97 & 89.94
& 46.50 & 77.80 & 86.40 \\
& & SpecPrefetch
& \best{48.83} & \best{85.60} & \best{92.31}
& 48.36 & 84.35 & 92.47 \\

\cmidrule(lr){2-9}

& \textbf{HumanEval}
& FATE
& 47.22 & 81.16 & 88.11
& 47.43 & \best{81.51} & 90.14 \\
& & Draft Model
& \NA & \NA & \NA
& 42.68 & 70.14 & 80.13 \\
& & ProMoE
& 47.30 & 79.62 & 86.69
& 41.50 & 66.70 & 75.50 \\
& & SpecPrefetch
& \best{48.68} & \best{85.48} & \best{92.16}
& \best{47.61} & \best{81.51} & \best{90.65} \\

\midrule

\multirow{12}{*}{\shortstack[c]{\textbf{VLM}\\\textbf{Workloads}}}
& \textbf{OCRBench}
& FATE
& 48.72 & 84.21 & 92.05
& 48.59 & 84.84 & 93.07 \\
& & Draft Model
& 48.51 & 84.54 & 93.01
& 47.83 & 83.39 & 91.36 \\
& & ProMoE
& 49.06 & 87.93 & 94.01
& 48.80 & 86.90 & 94.20 \\
& & SpecPrefetch
& \best{49.43} & \best{89.29} & \best{95.29}
& \best{49.36} & \best{89.29} & \best{96.45} \\

\cmidrule(lr){2-9}

& \textbf{ChartQA-Test}
& FATE
& 48.47 & 83.88 & 90.85
& 48.43 & 83.70 & 91.99 \\
& & Draft Model
& 49.04 & 85.88 & 92.96
& 48.53 & 83.76 & 92.27 \\
& & ProMoE
& 49.48 & 89.01 & 95.21
& 49.50 & 87.60 & 95.30 \\
& & SpecPrefetch
& \best{49.51} & \best{89.40} & \best{95.48}
& \best{49.63} & \best{89.61} & \best{96.83} \\

\cmidrule(lr){2-9}

& \textbf{HallusionBench}
& FATE
& 48.95 & 85.01 & 92.80
& 49.13 & 86.21 & 94.98 \\
& & Draft Model
& 48.98 & 85.61 & 93.36
& 49.18 & 87.33 & 95.19 \\
& & ProMoE
& 49.37 & 88.14 & 94.38
& 48.80 & 86.30 & 93.80 \\
& & SpecPrefetch
& \best{49.49} & \best{89.33} & \best{95.34}
& \best{49.25} & \best{88.72} & \best{95.96} \\

\bottomrule
\end{tabular}
}
\par
\vspace{2pt}
\begin{minipage}{\textwidth}
\footnotesize
\(R@M\) denotes ExpertRecall@\(M\) with \(M\) predicted candidates. 
The evaluated budgets are \(\{K/2,K,K+2\}\), where \(K=8\) for Qwen3-VL-30B-A3B and \(K=6\) for DeepSeek-VL2-Tiny. 
The maximum \(R@4\) and \(R@3\) are \(50\%\).
\end{minipage}
\end{table*}

\section{Experiment}
\subsection{Experimental Setup}
\label{sec:experimental_setup}

\paragraph{Models, Training Data, and Benchmarks.}
We evaluate SpecPrefetch on Qwen3-VL-30B-A3B and DeepSeek-VL2-Tiny, covering diverse sparse MoE architectures and routing behaviors~\citep{qwen3vl,deepseekvl2}. Predictors are trained on a heterogeneous multimodal instruction corpus, while evaluation is performed on disjoint VLM and LLM benchmarks. Detailed training data and benchmark settings are provided in the supplementary material.

\paragraph{Baselines.}
We compare SpecPrefetch with FATE, Draft Model, and ProMoE.
FATE exploits adjacent layer routing regularity without training,
while Draft Model and ProMoE use learned predictors for next
layer expert demand. All methods use matched candidate budgets
and the same transfer only protocol. Predicted experts are used
only for prefetching, while the frozen native router independently
determines the executed experts.

\paragraph{Evaluation Metrics.}
We report ExpertRecall@\(M\) for model-level prediction quality and
ReadyRecall for system-level expert availability. For each valid
token \((b,s)\in\Omega\), \(\mathcal{T}_{b,s}^{l+1}\) is the
native Top-\(K\) expert set, \(\mathcal{C}_{M}^{l+1}\) is the
Top-\(M\) predicted candidate set, and
\(\mathcal{A}^{l+1}(t)\) is the set of experts available on the
accelerator at time \(t\). The metrics are

\[
\operatorname{ExpertRecall}@M
=
\frac{1}{|\Omega|}
\sum_{(b,s)\in\Omega}
\frac{
\left|
\mathcal{T}_{b,s}^{l+1}
\cap
\mathcal{C}_{M}^{l+1}
\right|
}{K}.
\]

\[
\operatorname{ReadyRecall}
=
\frac{1}{|\Omega|}
\sum_{(b,s)\in\Omega}
\frac{
\left|
\mathcal{T}_{b,s}^{l+1}
\cap
\mathcal{A}^{l+1}
\left(
t_{l+1}^{\mathrm{exec}}
\right)
\right|
}{K}.
\]

We additionally report trainable predictor parameters and
decoding throughput in tokens per second.

\begin{table*}[t]
\centering
\caption{
Predictor design ablation across all workloads and candidate
budgets, together with trainable parameter comparison.
}
\label{tab:predictor_ablation_full}
\setlength{\tabcolsep}{3pt}
\renewcommand{\arraystretch}{1.08}
\scriptsize

\begin{minipage}[t]{0.49\textwidth}
\centering
\textbf{(a) Qwen3-VL-30B-A3B}

\vspace{2pt}
\resizebox{\linewidth}{0.094\textheight}{
\begin{tabular}{lccc@{\hspace{5pt}}ccc}
\toprule
& \multicolumn{3}{c}{\textbf{Single MLP}}
& \multicolumn{3}{c}{\textbf{SpecPrefetch}} \\
\cmidrule(lr){2-4}
\cmidrule(lr){5-7}
\textbf{Benchmark}
& R@4$^{\dagger}$ & R@8 & R@10
& R@4$^{\dagger}$ & R@8 & R@10 \\
\midrule
GSM8K
& 47.56 & 79.36 & 86.29
& \best{48.83} & \best{85.60} & \best{92.31} \\

HumanEval
& 44.77 & 71.98 & 78.63
& \best{48.68} & \best{85.48} & \best{92.16} \\

OCRBench
& 48.23 & 85.35 & 91.52
& \best{49.43} & \best{89.29} & \best{95.29} \\

ChartQA-Test
& 49.31 & 87.76 & 94.20
& \best{49.51} & \best{89.40} & \best{95.48} \\

HallusionBench
& 48.84 & 85.66 & 92.03
& \best{49.49} & \best{89.33} & \best{95.34} \\
\bottomrule
\end{tabular}
}
\end{minipage}
\hfill
\begin{minipage}[t]{0.49\textwidth}
\centering
\textbf{(b) DeepSeek-VL2-Tiny}

\vspace{2pt}
\resizebox{\linewidth}{0.094\textheight}{
\begin{tabular}{lccc@{\hspace{5pt}}ccc}
\toprule
& \multicolumn{3}{c}{\textbf{Single MLP}}
& \multicolumn{3}{c}{\textbf{SpecPrefetch}} \\
\cmidrule(lr){2-4}
\cmidrule(lr){5-7}
\textbf{Benchmark}
& R@3$^{\dagger}$ & R@6 & R@8
& R@3$^{\dagger}$ & R@6 & R@8 \\
\midrule
GSM8K
& 39.90 & 64.30 & 73.30
& \best{48.36} & \best{84.35} & \best{92.47} \\

HumanEval
& 33.70 & 53.40 & 61.60
& \best{47.61} & \best{81.51} & \best{90.65} \\

OCRBench
& 46.30 & 78.70 & 87.40
& \best{49.36} & \best{89.29} & \best{96.45} \\

ChartQA-Test
& 47.20 & 79.10 & 88.20
& \best{49.63} & \best{89.61} & \best{96.83} \\

HallusionBench
& 46.40 & 78.70 & 87.40
& \best{49.25} & \best{88.72} & \best{95.96} \\
\bottomrule
\end{tabular}
}
\end{minipage}

\vspace{5pt}

{\footnotesize
\(R@M\) denotes ExpertRecall@\(M\) with \(M\) predicted candidates.
\(R@4\) and \(R@3\) have a maximum possible value of 50\%.
}
\end{table*}

\begin{table}[t]
\centering
\caption{
Trainable parameters of different expert predictors.
}
\label{tab:trainable_params}
\setlength{\tabcolsep}{2pt}
\renewcommand{\arraystretch}{1.08}
\footnotesize
\begin{tabular*}{\columnwidth}{@{\extracolsep{\fill}}lcccc@{}}
\toprule
\multicolumn{1}{c}{\multirow{2}{*}{\textbf{Model}}}
& \textbf{Single}
& \textbf{Draft}
& \multirow{2}{*}{\textbf{ProMoE}}
& \multirow{2}{*}{\textbf{SpecPrefetch}} \\
& \textbf{MLP}
& \textbf{Model}
& & \\
\midrule
DeepSeek-VL2-Tiny
& \textbf{0.78M}
& 26.40M
& 16.35M
& 0.82M \\
Qwen3-VL-30B-A3B
& 12.19M
& 19.19M
& 207.23M
& \textbf{6.48M} \\
\bottomrule
\end{tabular*}
\end{table}

\paragraph{On Device Evaluation.}
We deploy the 4 bit DeepSeek-VL2-Tiny model on a Qualcomm
Snapdragon 8 Elite platform with an Adreno 825 GPU. The model
contains 11 MoE layers, each with 64 routed experts and native
Top-\(K\) routing with \(K=6\), and each quantized expert occupies
1.94 MB. We compare load on-demand execution, a compute
optimized runtime, and SpecPrefetch built on the same runtime.
Experiments use physical NVMe storage and emulate slower storage through controlled loading delays.

\paragraph{Implementation Details.}
The predictors are trained for one epoch using AdamW with
\(\beta=(0.9,0.999)\), \(\epsilon=10^{-8}\), and learning rate
\(10^{-4}\). We use cosine scheduling, a warmup ratio of 0.03,
gradient clipping at 0.5, and bf16 precision. The maximum input
length is 4096 for DeepSeek-VL2-Tiny and 2048 for
Qwen3-VL-30B-A3B. Predictor training uses PyTorch, HuggingFace Trainer, DeepSpeed, and FlashAttention 2
on \(2\times8\) NVIDIA A800-SXM4-80GB GPUs.

\begin{table}[t]
\centering
\caption{Runtime overhead of prediction and prefetch control under
cold-cache NVMe. PRED0 executes the complete prediction and control
path but issues no expert transfers.}
\label{tab:overhead}
\setlength{\tabcolsep}{3pt}
\renewcommand{\arraystretch}{1.08}
\footnotesize
\begin{tabular*}{\columnwidth}{@{\extracolsep{\fill}}lcc@{}}
\toprule
\multicolumn{1}{c}{\multirow{2}{*}{\textbf{Configuration}}}
& \textbf{Throughput}
& \textbf{Gain vs. OFF} \\
& \textbf{(tok/s)}
& \textbf{(\%)} \\
\midrule
OFF & \(3.753 \pm 0.164\) & -- \\
PRED0 & \(3.730 \pm 0.113\) & \(-0.53 \pm 3.05\) \\
SpecPrefetch (\(M_{\max}=8\)) & \(4.055 \pm 0.129\) & \(+8.12 \pm 2.79\) \\
\bottomrule
\end{tabular*}
\end{table}

\subsection{Comparison Results}

Table~\ref{tab:prediction_next1_comparison} compares SpecPrefetch with FATE, Draft Model, and ProMoE under a unified next-layer prediction protocol. R@\(M\) denotes ExpertRecall@\(M\), where \(M\) is the number of predicted expert candidates.

SpecPrefetch achieves the best or tied-best recall in 27 of the 30 settings. It ranks first across all settings on Qwen3-VL-30B-A3B and across HumanEval and all VLM benchmarks on DeepSeek-VL2-Tiny. Its largest gains over FATE are \(5.52\) percentage points at R@8 on ChartQA-Test with Qwen3-VL-30B-A3B and \(5.91\) percentage points at R@6 on ChartQA-Test with DeepSeek-VL2-Tiny. The only exceptions are the three DeepSeek-VL2-Tiny GSM8K settings, where FATE leads by \(0.26\), \(0.31\), and \(0.60\) percentage points.

SpecPrefetch also consistently outperforms the learned predictor baselines, with particularly clear advantages on VLM workloads. These results demonstrate reliable next-layer expert coverage across different model architectures, workloads, and candidate budgets.

\begin{table}[t]
\centering
\caption{Matched-budget comparison of expert prediction strategies on Snapdragon 8 Elite with \(M_{\max}=8\).}
\label{tab:window_budget}
\setlength{\tabcolsep}{1pt}
\renewcommand{\arraystretch}{1.08}
\footnotesize
\begin{tabular*}{\columnwidth}{@{\extracolsep{\fill}}lccc@{}}
\toprule
\multicolumn{1}{c}{\multirow{2}{*}{\textbf{Policy}}}
& \textbf{Cold NVMe}
& \textbf{Slow Storage}
& \textbf{ReadyRecall} \\
& \textbf{Gain (\%)}
& \textbf{Gain (\%)}
& \textbf{(\%)} \\
\midrule

Naive Prefetch
& \(-1.08 \pm 2.83\)
& \(-2.90 \pm 2.66\)
& \(23.4 \pm 7.3\) \\

SpecPrefetch

& \(\mathbf{6.97 \pm 2.56}\)
& \(\mathbf{15.43 \pm 2.96}\)
& \(\mathbf{91.7 \pm 3.1}\) \\
\bottomrule
\end{tabular*}
\end{table}

\subsection{Predictor Design and Parameter Efficiency}

Table~\ref{tab:predictor_ablation_full} compares SpecPrefetch with a Single MLP under identical supervision and candidate budgets. SpecPrefetch outperforms the Single MLP in all 30 settings, with particularly large gains on DeepSeek-VL2-Tiny. At the native candidate budget R@6, it improves recall by \(20.05\) and \(28.11\) percentage points on GSM8K and HumanEval, respectively, and by more than \(10\) percentage points on all three VLM benchmarks. The consistent gains across both models indicate that layer-specific predictors capture cross-layer routing transitions more effectively than a shared predictor.

As shown in Table~\ref{tab:trainable_params}, these improvements are achieved with limited trainable overhead. On Qwen3-VL-30B-A3B, SpecPrefetch uses \(6.48\)M parameters, approximately half the size of the Single MLP and only \(3.1\%\) of ProMoE. On DeepSeek-VL2-Tiny, its \(0.82\)M parameters are comparable to the \(0.78\)M Single MLP and correspond to only \(3.1\%\) and \(5.0\%\) of Draft Model and ProMoE, respectively. These results demonstrate that the layer-specific low-rank design improves expert coverage without relying on a substantially larger predictor.

\subsection{On-Device System Evaluation}
\label{sec:on_device_evaluation}

We evaluate whether improved expert prediction translates into practical decoding acceleration on a Snapdragon 8 Elite platform. The experiments use physical NVMe storage, together with controlled loading delays to emulate slower storage conditions. Unless otherwise specified, the main throughput comparisons use controlled paired measurements against the compute-optimized offloading runtime.

\begin{figure}[!t]
  \centering
  \includegraphics[width=\columnwidth]{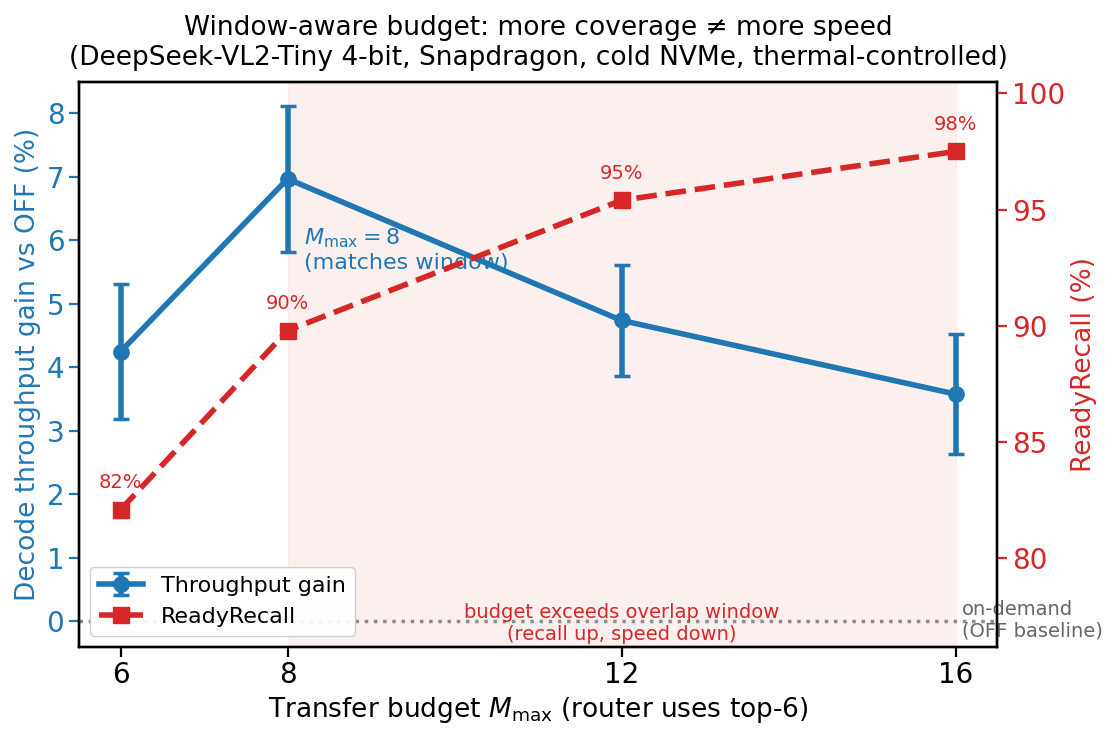}
\caption{
Impact of the maximum prefetch budget under cold-cache NVMe.
Throughput peaks at \(M_{\max}=8\), supporting bounded prefetching under the overlap window.
}
  \label{fig:res_realdevice}
\end{figure}

\begin{figure}[t]
  \centering
  \includegraphics[width=\columnwidth]{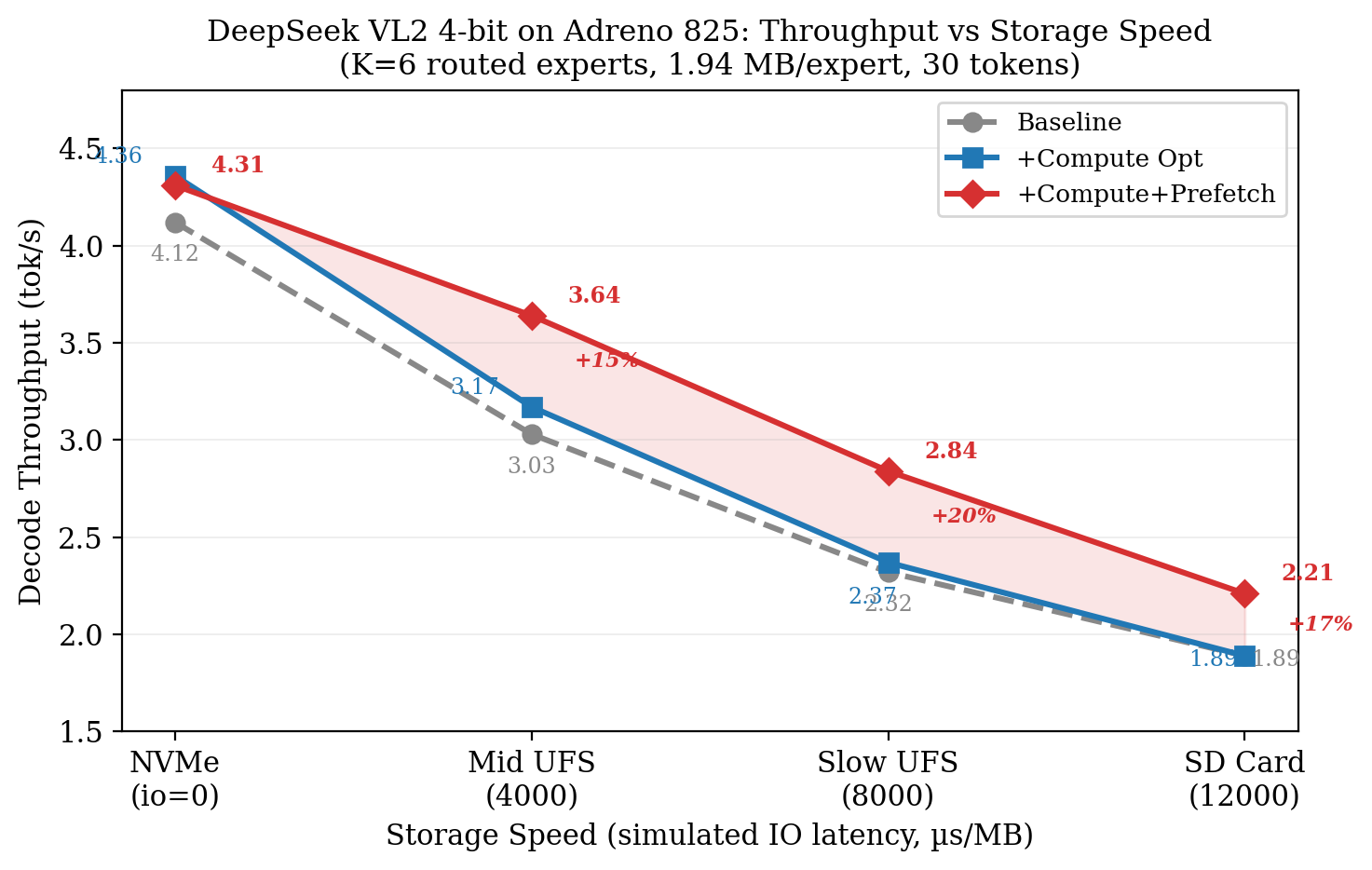}
  \caption{Storage sensitivity of on device decoding throughput under
physical NVMe and emulated loading delays. The benefit of
SpecPrefetch increases as expert loading becomes more exposed
on the inference critical path.}
  \label{fig:fig_toks}
\end{figure}

\paragraph{Prediction Overhead and End-to-End Benefit.}
Table~\ref{tab:overhead} decomposes the runtime effect of prediction-guided prefetching under cold-cache NVMe. OFF disables both prediction and prefetching, while PRED$_0$ executes the complete prediction and control path without issuing expert transfers. PRED$_0$ achieves \(3.730 \pm 0.113\) tokens/s, compared with \(3.753 \pm 0.164\) tokens/s for OFF, corresponding to a measured difference of only \(-0.53\%\). This indicates that the prediction and control path introduces limited runtime overhead. When asynchronous expert transfers are enabled, SpecPrefetch achieves \(4.055 \pm 0.129\) tokens/s, yielding an \(8.12\%\) improvement over OFF. These results show that the latency hidden by expert prefetching outweighs the additional prediction and control overhead.

\paragraph{Prediction Quality under a Matched Prefetch Budget.}
Table~\ref{tab:window_budget} compares SpecPrefetch with Naive Prefetch under the same maximum prefetch budget of \(M_{\max}=8\). Naive Prefetch uses the experts activated by the preceding token, whereas SpecPrefetch selects candidates according to the learned expert priorities. SpecPrefetch increases ReadyRecall from \(23.4\%\) to \(91.7\%\), resulting in throughput gains of \(6.97\%\) under cold NVMe and \(15.43\%\) under slow storage. In contrast, Naive Prefetch provides no acceleration and introduces additional overhead. These results show that accurate prediction translates a limited prefetch budget into higher expert readiness and throughput.

\paragraph{Prefetch Budget Sensitivity.}
Figure~\ref{fig:res_realdevice} examines the effect of \(M_{\max}\) under physical cold-cache NVMe. Increasing the budget improves ReadyRecall from \(82.1\%\) to \(97.5\%\), but throughput peaks at \(M_{\max}=8\) and decreases with larger budgets. This shows that higher expert coverage does not necessarily yield greater acceleration, since excessive prefetching introduces redundant transfers and additional I/O pressure. It therefore supports using a bounded prefetch budget determined by the computation and transfer overlap window.

\paragraph{Storage Sensitivity.}
Figure~\ref{fig:fig_toks} evaluates decoding throughput under physical NVMe and emulated storage delays. SpecPrefetch provides limited additional benefit when expert loading is largely hidden under fast NVMe, but its advantage increases as storage latency becomes more exposed on the inference critical path. Compared with the compute-optimized runtime, the throughput improvement reaches approximately \(20\%\) under the Slow UFS setting and \(17.02\%\) under the slowest SD-card setting. The experiment evaluates sensitivity to storage latency rather than shared bandwidth contention.

\begin{figure}[t]
  \centering
  \includegraphics[width=0.9\columnwidth]{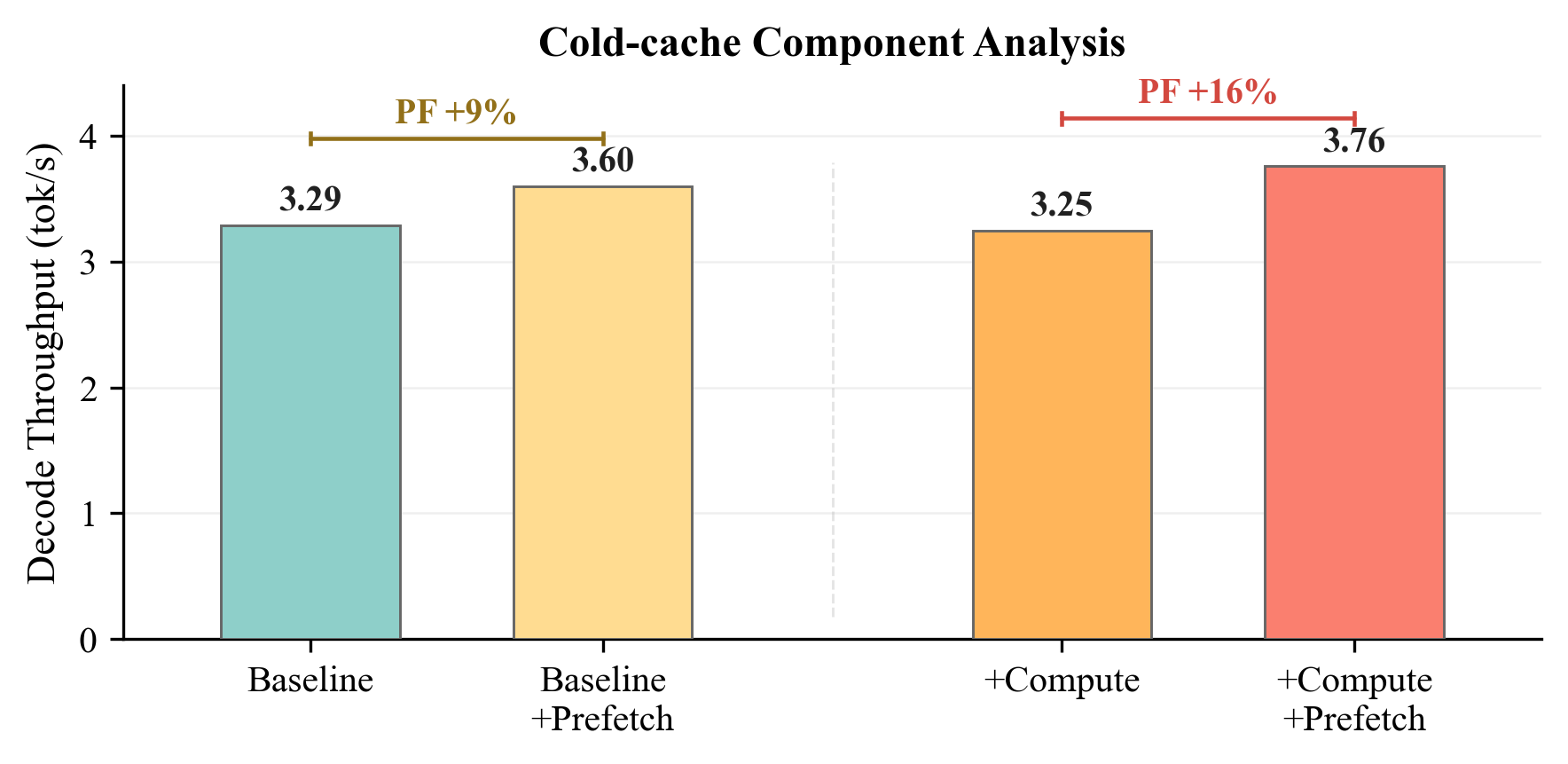}
  \caption{Cold cache component analysis on Snapdragon 8 Elite after
clearing the operating system page cache. The comparison
illustrates the complementary roles of compute optimization and
prediction guided expert prefetching.}
  \label{fig:fig_cold_cache}
\end{figure}

\paragraph{Complementarity with Compute Optimization.}
Figure~\ref{fig:fig_cold_cache} compares prediction-guided prefetching with compute optimization after clearing the operating system page cache. Their combination achieves the highest throughput, indicating that the two techniques address complementary bottlenecks: compute optimization reduces execution and dispatch overhead, while SpecPrefetch hides exposed expert-loading latency. Because mobile throughput is sensitive to thermal state, this experiment is used as a component analysis rather than as the primary quantitative comparison.


\section{Conclusion}

We present SpecPrefetch, a router-preserving expert prefetching framework for efficient sparse MoE inference. By predicting next-layer expert priorities with lightweight adapters and performing window-aware asynchronous transfers, SpecPrefetch improves expert availability without modifying native routing decisions. Experiments on representative MoE models and real-device deployment demonstrate consistent gains in inference efficiency, validating the effectiveness of prediction-guided and budget-aware expert prefetching for practical MoE deployment.

\bibliography{aaai2027}

@misc{qwen3vl,
  title = {{Qwen3-VL} Technical Report},
  author = {{Qwen Team}},
  year = {2025},
  eprint = {2511.21631},
  archivePrefix = {arXiv},
  primaryClass = {cs.CV}
}

@misc{deepseekvl2,
  title = {{DeepSeek-VL2}: Mixture-of-Experts Vision-Language Models for Advanced Multimodal Understanding},
  author = {Wu, Zhiyu and Chen, Xiaokang and Pan, Zizheng and Liu, Xingchao and Liu, Wen and Dai, Damai and Gao, Huazuo and Ma, Yiyang and Wu, Chengyue and Wang, Bingxuan and others},
  year = {2024},
  eprint = {2412.10302},
  archivePrefix = {arXiv},
  primaryClass = {cs.CV}
}

@article{ocrbench,
  title = {{OCRBench}: On the Hidden Mystery of {OCR} in Large Multimodal Models},
  author = {Liu, Yuliang and Li, Zhang and Huang, Mingxin and Yang, Biao and Yu, Wenwen and Li, Chunyuan and Yin, Xucheng and Liu, Cheng-lin and Jin, Lianwen and Bai, Xiang},
  journal = {Science China Information Sciences},
  year = {2024},
  eprint = {2305.07895},
  archivePrefix = {arXiv},
  doi = {10.1007/s11432-024-4235-6}
}

@inproceedings{hallusionbench,
  title = {{HallusionBench}: An Advanced Diagnostic Suite for Entangled Language Hallucination and Visual Illusion in Large Vision-Language Models},
  author = {Guan, Tianrui and Liu, Fuxiao and Wu, Xiyang and Xian, Ruiqi and Li, Zongxia and Liu, Xiaoyu and Wang, Xijun and Chen, Lichang and Huang, Furong and Yacoob, Yaser and Manocha, Dinesh and Zhou, Tianyi},
  booktitle = {Proceedings of the IEEE/CVF Conference on Computer Vision and Pattern Recognition},
  year = {2024}
}

@misc{gsm8k,
  title = {Training Verifiers to Solve Math Word Problems},
  author = {Cobbe, Karl and Kosaraju, Vineet and Bavarian, Mohammad and Chen, Mark and Jun, Heewoo and Kaiser, Lukasz and Plappert, Matthias and Tworek, Jerry and Hilton, Jacob and Nakano, Reiichiro and Hesse, Christopher and Schulman, John},
  year = {2021},
  eprint = {2110.14168},
  archivePrefix = {arXiv},
  primaryClass = {cs.LG}
}

@misc{humaneval,
  title = {Evaluating Large Language Models Trained on Code},
  author = {Chen, Mark and Tworek, Jerry and Jun, Heewoo and Yuan, Qiming and Pinto, Henrique Ponde de Oliveira and Kaplan, Jared and Edwards, Harri and Burda, Yuri and Joseph, Nicholas and Brockman, Greg and others},
  year = {2021},
  eprint = {2107.03374},
  archivePrefix = {arXiv},
  primaryClass = {cs.LG}
}

@misc{qwen_moe,
  title = {{Qwen3 Technical Report}},
  author = {{Qwen Team}},
  year = {2025},
  eprint = {2505.09388},
  archivePrefix = {arXiv},
  primaryClass = {cs.CL},
  url = {https://arxiv.org/abs/2505.09388}
}

@misc{moepic2025,
  title = {Accelerating Mixture-of-Expert Inference with Adaptive Expert Split Mechanism},
  author = {Yan, Jiaming and Liu, Jianchun and Xu, Hongli and Huang, Liusheng},
  year = {2025},
  eprint = {2509.08342},
  archivePrefix = {arXiv},
  primaryClass = {cs.LG},
  url = {https://arxiv.org/abs/2509.08342}
}

@misc{fang2025fate,
  title = {{FATE}: Fast Edge Inference of Mixture-of-Experts Models via Cross-Layer Gate Reuse},
  author = {Fang, Zhiyuan and Yang, Tianle and Wang, Yang and Xu, Jilong and Liu, Yifan and Chen, Beidi and Xing, Eric and Zhang, Zhihao and Chen, Yiran},
  year = {2025},
  eprint = {2502.12224},
  archivePrefix = {arXiv},
  primaryClass = {cs.LG},
  url = {https://arxiv.org/abs/2502.12224}
}

@inproceedings{hwang2024pregated,
  title = {Pre-gated {MoE}: An Algorithm-System Co-Design for Fast and Scalable Mixture-of-Expert Inference},
  author = {Hwang, Ranggi and Wei, Jianyu and Cao, Shijie and Hwang, Changho and Tang, Xiaohu and Cao, Ting and Yang, Mao},
  booktitle = {Proceedings of the 51st Annual International Symposium on Computer Architecture},
  year = {2024}
}

@misc{chen2025spmoe,
  title = {{SPMoE}: Accelerating Sparse Mixture-of-Experts Inference with Learned Expert Prediction},
  author = {Chen, Yujun and Li, Chao and Zhang, Mingxuan and Wang, Wenfeng and Liu, Jiacheng and Guo, Minyi},
  year = {2025},
  note = {Preprint}
}

@inproceedings{leviathan2023speculative,
  title = {Fast Inference from Transformers via Speculative Decoding},
  author = {Leviathan, Yaniv and Kalman, Matan and Matias, Yossi},
  booktitle = {Proceedings of the 40th International Conference on Machine Learning},
  year = {2023}
}

@misc{eagle2024,
  title = {{EAGLE}: Speculative Sampling Requires Rethinking Feature Uncertainty},
  author = {Li, Yuhui and Wei, Fangyun and Zhang, Chao and Zhang, Hongyang},
  year = {2024},
  eprint = {2401.15077},
  archivePrefix = {arXiv},
  primaryClass = {cs.CL},
  url = {https://arxiv.org/abs/2401.15077}
}

@inproceedings{shazeer2017outrageously,
  title = {Outrageously Large Neural Networks: The Sparsely-Gated Mixture-of-Experts Layer},
  author = {Shazeer, Noam and Mirhoseini, Azalia and Maziarz, Krzysztof and Davis, Andy and Le, Quoc and Hinton, Geoffrey and Dean, Jeff},
  booktitle = {International Conference on Learning Representations},
  year = {2017}
}

@inproceedings{lepikhin2020gshard,
  title = {{GShard}: Scaling Giant Models with Conditional Computation and Automatic Sharding},
  author = {Lepikhin, Dmitry and Lee, HyoukJoong and Xu, Yuanzhong and Chen, Dehao and Firat, Orhan and Huang, Yanping and Krikun, Maxim and Shazeer, Noam and Chen, Zhifeng},
  booktitle = {International Conference on Learning Representations},
  year = {2021}
}

@article{fedus2022switch,
  title = {Switch Transformers: Scaling to Trillion Parameter Models with Simple and Efficient Sparsity},
  author = {Fedus, William and Zoph, Barret and Shazeer, Noam},
  journal = {Journal of Machine Learning Research},
  volume = {23},
  number = {120},
  pages = {1--39},
  year = {2022}
}

@inproceedings{du2022glam,
  title = {{GLaM}: Efficient Scaling of Language Models with Mixture-of-Experts},
  author = {Du, Nan and Huang, Yanping and Dai, Andrew M. and Tong, Simon and Lepikhin, Dmitry and Xu, Yuanzhong and Krikun, Maxim and Zhou, Yanqi and Yu, Adams Wei and Firat, Orhan and Zoph, Barret and Fedus, Liam and Bosma, Maarten and Zhou, Zongwei and Wang, Tao and Wang, Yu Emma and Webster, Kellie and Pellat, Marie and Robinson, Kevin and Meier-Hellstern, Kathleen and Duke, Toju and Dixon, Lucas and Zhang, Kun and Le, Quoc V. and Wu, Yonghui and Chen, Zhifeng and Cui, Claire},
  booktitle = {Proceedings of the 39th International Conference on Machine Learning},
  year = {2022}
}

@misc{jiang2024mixtral,
  title = {{Mixtral of Experts}},
  author = {Jiang, Albert Q. and Sablayrolles, Alexandre and Roux, Antoine and Mensch, Arthur and Savary, Blanche and Bamford, Chris and Chaplot, Devendra Singh and de las Casas, Diego and Hanna, Emma Bou and Bressand, Florian and Lengyel, Gianna and Bour, Guillaume and Lample, Guillaume and Lavaud, L{\'e}lio Renard and Saulnier, Lucile and Lachaux, Marie-Anne and Stock, Pierre and Le Scao, Teven and Lavril, Thibaut and Wang, Thomas and Lacroix, Timoth{\'e}e and El Sayed, William},
  year = {2024},
  eprint = {2401.04088},
  archivePrefix = {arXiv},
  primaryClass = {cs.LG},
  url = {https://arxiv.org/abs/2401.04088}
}

@misc{dai2024deepseekmoe,
  title = {{DeepSeekMoE}: Towards Ultimate Expert Specialization in Mixture-of-Experts Language Models},
  author = {Dai, Damai and Deng, Chengqi and Zhao, Chenggang and Xu, R. X. and Gao, Huazuo and Chen, Deli and Li, Jiashi and Zeng, Wangding and Yu, Xingkai and Wu, Y. and Xie, Zhenda and Li, Y. K. and Huang, Panpan and Luo, Fuli and Ruan, Chong and Sui, Zhifang and Liang, Wenfeng},
  year = {2024},
  eprint = {2401.06066},
  archivePrefix = {arXiv},
  primaryClass = {cs.CL},
  url = {https://arxiv.org/abs/2401.06066}
}

@misc{komatsuzaki2023sparse,
  title = {Sparse Upcycling: Training Mixture-of-Experts from Dense Checkpoints},
  author = {Komatsuzaki, Aran and Puigcerver, Joan and Lee-Thorp, James and Ruiz, Carlos Riquelme and Mustafa, Basil and Ainslie, Joshua and Tay, Yi and Dehghani, Mostafa and Houlsby, Neil},
  year = {2023},
  eprint = {2212.05055},
  archivePrefix = {arXiv},
  primaryClass = {cs.LG},
  url = {https://arxiv.org/abs/2212.05055}
}

@misc{semoe2022,
  title = {{SE-MoE}: A Scalable and Efficient Mixture-of-Experts Distributed Training and Inference System},
  author = {Shen, Liang and Wu, Zhihua and Gong, Weibao and Hao, Hongxiang and Bai, Yangfan and Wu, Huachao and Wu, Xinxuan and Bian, Jiang and Xiong, Haoyi and Yu, Dianhai and Ma, Yanjun},
  year = {2022},
  eprint = {2205.10034},
  archivePrefix = {arXiv},
  primaryClass = {cs.DC},
  url = {https://arxiv.org/abs/2205.10034}
}

@misc{moeinfinity2025,
  title = {{MoE-Infinity}: Efficient {MoE} Inference on Personal Machines with Sparsity-Aware Expert Cache},
  author = {Xue, Leyang and Fu, Yao and Lu, Zhan and Sun, Chuanhao and Mai, Luo and Marina, Mahesh},
  year = {2024},
  eprint = {2401.14361},
  archivePrefix = {arXiv},
  primaryClass = {cs.LG},
  url = {https://arxiv.org/abs/2401.14361}
}

@misc{expertflow2024,
  title = {{ExpertFlow}: Efficient Mixture-of-Experts Inference via Predictive Expert Caching and Token Scheduling},
  author = {He, Xin and Zhang, Shunkang and Tang, Kaijie and Shi, Shaohuai and Wang, Yuxin and Zeng, Zihao and Tang, Zhenheng and Chu, Xiaowen and Yin, Haiyan and Tsang, Ivor W. and Ong, Yew Soon},
  year = {2024},
  eprint = {2410.17954},
  archivePrefix = {arXiv},
  primaryClass = {cs.AI},
  url = {https://arxiv.org/abs/2410.17954}
}

@misc{layerscope2024,
  title = {{LayerScope}: Predictive Cross-Layer Scheduling for Efficient Multi-Batch {MoE} Inference on Legacy Servers},
  author = {Yu, Enda and Dong, Dezun and Zhang, Zhaoning and Bai, Zhe and Yang, Weiling and Wang, Haojie and Li, Dongsheng and Wu, Yongwei and Liao, Xiangke},
  year = {2025},
  eprint = {2509.23638},
  archivePrefix = {arXiv},
  primaryClass = {cs.LG},
  url = {https://arxiv.org/abs/2509.23638}
}

@misc{klotski2025,
  title = {{Klotski}: Efficient Mixture-of-Expert Inference via Expert-Aware Multi-Batch Pipeline},
  author = {Fang, Zhiyuan and Huang, Yuegui and Hong, Zicong and Lyu, Yufeng and Chen, Wuhui and Yu, Yue and Yu, Fan and Zheng, Zibin},
  year = {2025},
  eprint = {2502.06888},
  archivePrefix = {arXiv},
  primaryClass = {cs.DC},
  url = {https://arxiv.org/abs/2502.06888}
}

@misc{hobbit2024,
  title = {{HOBBIT}: A Mixed Precision Expert Offloading System for Fast {MoE} Inference},
  author = {Tang, Peng and Liu, Jiacheng and Hou, Xiaofeng and Pu, Yifei and Wang, Jing and Heng, Pheng-Ann and Li, Chao and Guo, Minyi},
  year = {2024},
  eprint = {2411.01433},
  archivePrefix = {arXiv},
  primaryClass = {cs.LG},
  url = {https://arxiv.org/abs/2411.01433}
}

@misc{finemoe2026,
  title = {Taming Latency-Memory Trade-Off in {MoE}-Based {LLM} Serving via Fine-Grained Expert Offloading},
  author = {Yu, Hanfei and Cui, Xingqi and Zhang, Hong and Wang, Hao and Wang, Hao},
  year = {2025},
  eprint = {2502.05370},
  archivePrefix = {arXiv},
  primaryClass = {cs.LG},
  url = {https://arxiv.org/abs/2502.05370}
}

@misc{moeeras2024,
  title = {{MoE-ERAS}: Efficient Runtime and Scheduling for Mixture-of-Experts Serving},
  author = {Abhimanyu Bambhaniya, Sashankh Chengavalli Kumar, Tushar Krishna},
  year = {2024},
  note = {Preprint}
}

@inproceedings{dsmoe2024,
  title = {{DS-MoE}: Dynamic Expert Scheduling for Efficient {MoE}-based {LLM} Inference},
  author = {Sun, Qi and Li, Yu},
  booktitle = {Proceedings of the 5th International Conference on Intelligent Technology and Embedded Systems},
  pages = {90--95},
  year = {2025},
  doi = {10.1109/ICITES66466.2025.11274315}
}

@misc{moebeyond2025,
  title = {{MoE-Beyond}: Learning-Based Expert Activation Prediction on Edge Devices},
  author = {Gavhane, Nishant and Mehrotra, Arush and Chawla, Rohit and Proenca, Peter},
  year = {2025},
  eprint = {2508.17137},
  archivePrefix = {arXiv},
  primaryClass = {cs.LG},
  url = {https://arxiv.org/abs/2508.17137}
}

@misc{preattention2025,
  title = {Pre-Attention Expert Prediction and Prefetching for Mixture-of-Experts Large Language Models},
  author = {Zhu, Shien and Bohl, Samuel and Oester, Robin and Alonso, Gustavo},
  year = {2025},
  eprint = {2511.10676},
  archivePrefix = {arXiv},
  primaryClass = {cs.CL},
  url = {https://arxiv.org/abs/2511.10676}
}

@misc{moespeq2025,
  title = {{MoE-SpeQ}: Speculative Quantized Decoding with Proactive Expert Prefetching and Offloading for Mixture-of-Experts},
  author = {Wang, Wenfeng and Liu, Jiacheng and Hou, Xiaofeng and Xia, Xinfeng and Tang, Peng and Zhang, Mingxuan and Li, Chao and Guo, Minyi},
  year = {2025},
  eprint = {2511.14102},
  archivePrefix = {arXiv},
  primaryClass = {cs.LG},
  url = {https://arxiv.org/abs/2511.14102}
}


\end{document}